\documentclass[pdflatex,sn-mathphys,Numbered]{sn-jnl}

\usepackage{soul}
\usepackage{graphicx}%
\usepackage{multirow}%
\usepackage{amsmath,amssymb,amsfonts}%
\usepackage{amsthm}%
\usepackage{mathrsfs}%
\usepackage[title]{appendix}%
\usepackage{xcolor}%
\usepackage{textcomp}%
\usepackage{manyfoot}%
\usepackage{booktabs}%
\usepackage{algorithm}%
\usepackage{algorithmicx}%
\usepackage{algpseudocode}%
\usepackage{listings}%
\usepackage{enumitem}%

\newcommand{\etal}{\textit{et al.}}
\DeclareMathOperator{\sign}{sign}

\begin{document}

\title[Article Title]{Body-mounted MR-conditional Robot for Minimally Invasive Liver Intervention}

\author[1]{\fnm{Zhefeng} \sur{Huang}}\email{zhuang480@gatech.edu}

\author[1]{\fnm{Anthony} \spfx{L.} \sur{Gunderman}}\email{agunderman3@gatech.edu}

\author[1]{\fnm{Samuel} \spfx{E.} \sur{Wilcox}}\email{swilcox33@gatech.edu}

\author[2]{\fnm{Saikat} \sur{Sengupta}}\email{saikat.sengupta@vumc.org}

\author[3]{\fnm{Jay} \sur{Shah}}\email{jay.shah@emory.edu}

\author[4]{\fnm{Aiming} \sur{Lu}}\email{Lu.Aiming@mayo.edu}

\author[4]{\fnm{David} \sur{Woodrum}}\email{Woodrum.David@mayo.edu}

\author*[1,5]{\fnm{Yue} \sur{Chen}}\email{yue.chen@bme.gatech.edu}

\affil[1]{\orgdiv{Institute of Robotics and Intelligent Machines}, \orgname{Georgia Institute of Technology}, \orgaddress{\street{801 Atlantic Dr NW}, \city{Atlanta}, \postcode{30332}, \state{GA}, \country{USA}}}

\affil[2]{\orgdiv{Vanderbilt University Institute of Imaging Science}, \orgname{Vanderbilt University Medical Center}, \orgaddress{\street{1161 21st Ave South Medical Center North}, \city{Nashville}, \postcode{37232}, \state{TN}, \country{USA}}}

\affil[3]{\orgdiv{Department of Radiology}, \orgname{Emory University}, \orgaddress{\street{1364 Clifton Rd}, \city{Atlanta}, \postcode{30329}, \state{GA}, \country{USA}}}

\affil[4]{\orgdiv{Department of Radiology}, \orgname{Mayo Clinic}, \orgaddress{\street{200 1st St SW}, \city{Rochester}, \postcode{55905}, \state{MN}, \country{USA}}}

\affil*[5]{\orgdiv{Department of Biomedical Engineering}, \orgname{Georgia Institute of Technology/Emory University}, \orgaddress{\street{313 Ferst Dr  2127}, \city{Atlanta}, \postcode{30332}, \state{GA}, \country{USA}}}

\abstract{MR-guided microwave ablation (MWA) has proven effective in treating hepatocellular carcinoma (HCC) with small-sized tumors, but the state-of-the-art technique suffers from sub-optimal workflow due to speed and accuracy of needle placement. This paper presents a compact body-mounted MR-conditional  robot that can operate in closed-bore MR scanners for accurate needle guidance. The robotic platform consists of two stacked Cartesian XY stages, each with two degrees of freedom, that facilitate needle guidance. The robot is actuated using 3D-printed pneumatic turbines with MR-conditional bevel gear transmission systems. Pneumatic valves and control mechatronics are located inside the MRI control room and are connected to the robot with pneumatic transmission lines and optical fibers. 
Free space experiments indicated robot-assisted needle insertion  error of 2.6±1.3 mm at an insertion depth of 80 mm. The MR-guided phantom studies were conducted to verify the MR-conditionality and targeting performance of the robot. Future work will focus on the system optimization and validations in animal trials. }

\keywords{Magnetic Resonance Imaging, Medical Robot, MR-guided Intervention}

\maketitle

\section{Introduction}\label{sec1}

Hepatocellular carcinoma (HCC) is the most common type of primary liver cancer and  stands as a preeminent contributor to cancer-associated mortalities across the world \cite{balogh2016hepatocellular, https://doi.org/10.3322/caac.21708}. The incidence rate of HCC continues to rise due to the increasing number of nonalcoholic fatty liver disease (NAFLD) and nonalcoholic steatohepatitis (NASH) \cite{ghouri2017review, mcglynn2021epidemiology}. 
A variety of methods can be applied to treat HCC, including resection, transplantation, percutaneous ethanol injection (PEI), radiofrequency ablation (RFA), microwave ablation (MWA), transarterial chemoembolization (TACE), transarterial radioembolization (TARE), systemic chemotherapy, multikinase or immune checkpoint inhibitors, etc. \cite{cabrera2010management, yang2020new}. Among these treatments, local ablations such as RFA and MWA are effective for treating early-stage HCC \cite{cabrera2010management, chen2020recent}. RFA and MWA offer effective curative outcomes in a minimally invasive manner, facilitating focal therapy treatment of the tumor while minimizing damage to surrounding healthy tissue. Compared to RFA, MWA has unique advantages including a wider active heating area and the capability for simultaneous activation of multiple antennae, enabling more rapid treatment of large or multifocal tumors. \cite{facciorusso2016local}. 

Despite being a widely accepted approach, accurate and efficient delivery of thermal energy poses a significant technical challenge due to difficulties involved with accurate MWA needle placement \cite{hoffmann2017mr}. Intraprocedural guidance, such as Ultrasound (US), Computed Tomography (CT) and Magnetic Resonance Imaging (MRI), is typically employed to improve targeting accuracy and to optimize needle position relative to the tumor \cite{kamarinos2020image}. Of these, MRI offers the best soft tissue resolution and best ability to visualize liver tumors. Additionally, MRI can provide (i) accurate 3D guidance towards the lesion \cite{gedroyc2005magnetic}, (ii) high-resolution soft tissue imaging, and (iii) thermometry feedback. However, a major limitation is that most MRI scanners are closed bore and provide limited access for needle insertion procedures \cite{hata2018robotics, monfaredi2018mri}. This difficulty has motivated the development of robot-assisted needle insertions. 

Many robots have been proposed to assist MR-guided (magnetic resonance imaging guided) needle insertions, including prostate ablations  \cite{chen2016robotic, chen2017mri}, neurosurgery \cite{li2014robotic, gunderman2022surgical, chen2019mr}, breast biopsy \cite{groenhuis2017stormram}, shoulder arthrography \cite{monfaredi2014prototype, monfaredi2018development}, and many more \cite{chen2019stereotactic}. Robotic systems utilized in the context of abdominal interventions are typically classified into two primary categories: (i) those that are affixed directly to the patient's body, denoted as body-mounted robots, and (ii) those that are fixed to either the MRI gantry, surgical table, or the surrounding floor, denoted as table-mounted robots  \cite{arnolli2015overview}. An early design of a table-mounted 3-degree of freedom (DoF) robot was developed to guide an ablation needle holder via the remote center-of-motion (RCM) mechanism \cite{hata2005needle}. However, to fully control the position and orientation of the straight needle, a total of 5-DoF is needed (or 4-DoF for needle guidance and 1-DoF for insertion that can be manually controlled). Christoforou \etal \cite{christoforou2014novel} proposed a table-mounted robotic mechanism with 5-DoF that is manually actuated. A fully automated MR-conditional and table-mounted robotic prototype endowed with 4-DoF was created by Franco \etal \cite{franco2015needle}. However, table-mounted robotic systems are subject to errors caused by tissue movement induced by physiological functions, such as breathing, or potential unexpected patient movement.

In addition to the table-mounted designs, many body-mounted mechanisms have also been developed. Body-mounted robotic systems potentially reduce the needle placement error caused by the reasons listed above by  allowing passive movement with the patient's body \cite{arnolli2015overview, musa2021respiratory, bricault2008light,gunderman2023autonomous}. A body-mounted double-ring mechanism was reported by Hata \etal \cite{hata2016body} for 2-DoF needle guidance. Bricault \etal   \cite{bricault2008light} proposed a body-mounted ``light puncture robot'' (LPR) that can actively manipulate 5-DoF of a needle within the MRI environment. Nevertheless, the support frame of LPR exhibits a relatively substantial volume, which could potentially conflict with the placement of body coils. A more compact   4-DoF body-mounted robot was developed by Li \etal{} using piezoelectric motors to control the position of two stacked Cartesian stages \cite{li2020body}. However, the use of piezoelectric motors often precludes image acquisition during robot motion \cite{xiao2020mr}. 

The primary challenges in interventional MR-guided abdominal robot development include: 1) ensuring MR-conditionality and MR imaging quality; 2) size constraints imposed by the closed MRI bore and body imaging coils; and 3) accurate control of the 4-DoF required  for effective needle guidance. In this paper a  compact, pneumatic motor driven, MR-conditional body-mounted robot is proposed for accurate needle guidance during MR-guided percutaneous interventions. Our main contributions include: 1) the design, manufacturing, and modeling of the robot hardware and 2) robot validations with free-space targeting and MRI phantom trials. 

\section{Materials and Methods}\label{sec2}

\subsection{Hardware Design and Fabrication}\label{subsec1}

In this work, we aim to develop a robotic platform that is capable of controlling the needle insertion vector within the MRI bore. Note that we propose a surgeon-in-loop approach for the insertion DoF. As such, the robot must possess the following necessary characteristics and capabilities. 1) The robot must have 4 actively controlled DoFs (that are not needle insertion or needle roll) to facilitate effective manipulation of the needle insertion vector. 2) The robot must be MR-conditional and all its components must be non-ferromagnetic to avoid interference with the magnetic field. 3) The robot design must ensure seamless employment with the MRI body coils and provide sufficient space for physicians to perform needle insertion.

\subsubsection{Robot Design}\label{subsubsec1}

\begin{figure}[t]
    \begin{center}
    \includegraphics[width=0.68\textwidth]{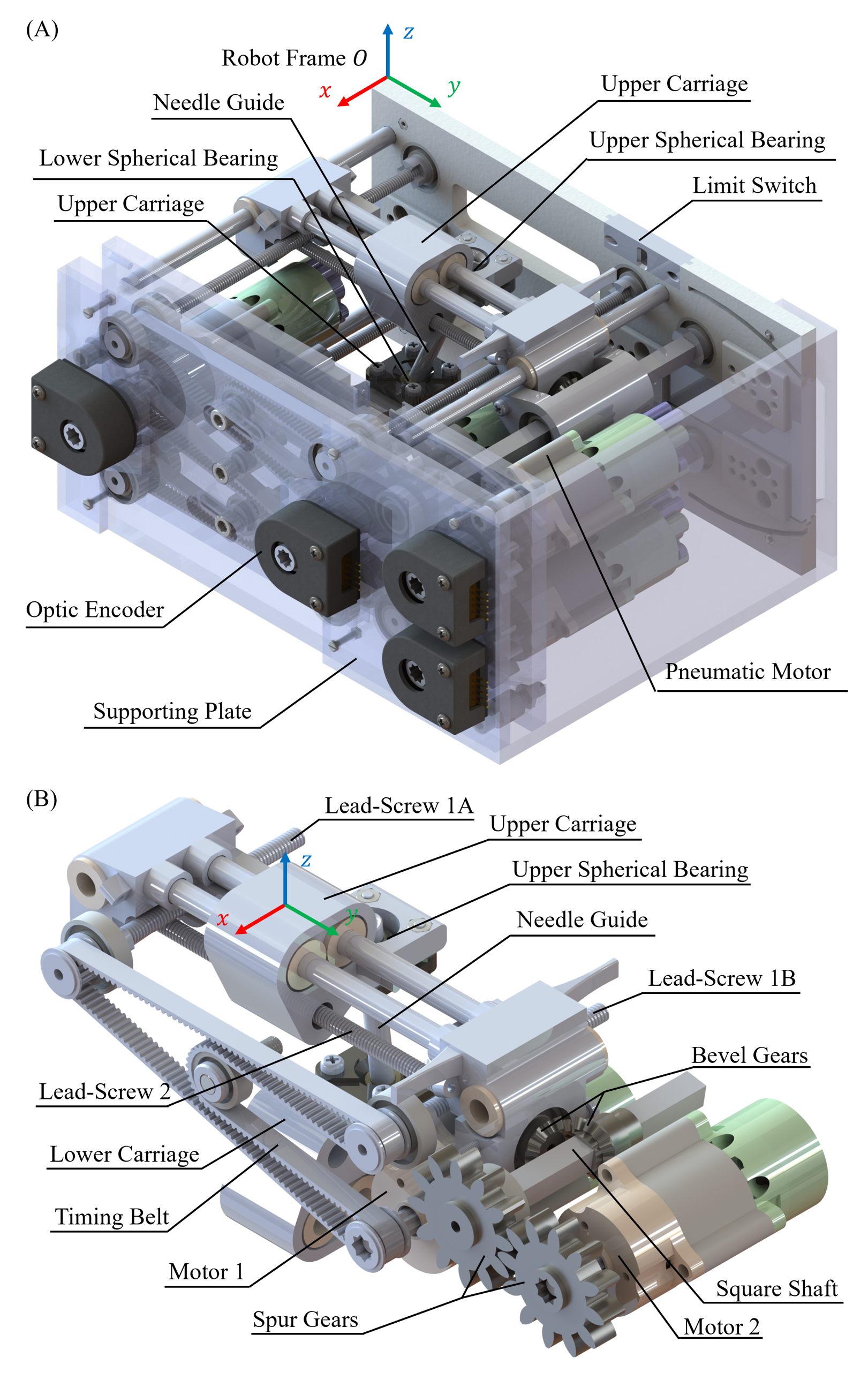}
    \end{center}
    \vspace{-2mm}
    \caption{(A) Overall robot design: The robot is composed of the following key components: supporting plates, transmission system, actuating system (including pneumatic motors, optic encoders and limit switches), and needle guide. (B) Upper motorized Cartesian stage transmission system design: The robot comprises two separate Cartesian stages for actuating the upper spherical bearing and the lower spherical bearing respectively. The components highlighted in this figure collectively constitute the transmission subsystem that drives the upper Cartesian stage.}
    \label{fig: RDO}
\end{figure}

The proposed robot, illustrated in Fig. \ref{fig: RDO}A, consists of two primary sub-systems: (i) a lower motorized Cartesian stage, and (ii) an upper motorized Cartesian stage. Each Cartesian stage possesses a carriage that has 2-DoF, providing linear translation motion capability in both the $x-$ and $y-$axis direction of the robot coordinate frame. The stage corresponding to displacements in the $x-$axis direction is referred to as the $x-$stage and the stage corresponding to displacements in the $y-$axis direction is referred to as the $y-$stage. Each carriage has embedded in it a spherical joint (EFSM-06, igus, Germany). As the carriages move relative to one another, a needle guide retained by the two spherical joints adjusts the needle insertion vector, providing 4 actively controlled DoFs for the needle pose. Note that the needle guide is solely connected to the upper carriage, and a relative sliding motion between the needle guide and the lower carriage is permitted. Feedback of the displacement of each carriage is obtained via MR-conditional encoders (EM2, US Digital, USA), attached to the $x-$ and $y-$ stage, and fiber optic limit switches. 

For the remainder of the section, we will consider the design of the upper Cartesian stage (Fig. \ref{fig: RDO}B), which is identical in operation to the lower Cartesian stage. Each axis stage of the upper Cartesian stage is actuated by a single, MR-safe pneumatic motor \cite{chen2017characterization,gunderman2023open}. And each major axis of the parts in Fig. \ref{fig: RDO}B is parallel to $x-$axis or $y-$axis of the robot coordinate frame attached at point $O$. The $x-$stage has two carriers that support the $y-$stage. The two carriers defining the $y-$stage's position are displaced using two translational lead-screws (lead-screw 1A and 1B in Fig. \ref{fig: RDO}B). These lead-screws are actuated by a single motor (motor 1) using a belt and pulley system, as depicted in Fig. \ref{fig: RDO}B. The $y-$stage is responsible for directly displacing the carriage retaining the spherical joint. This displacement is also performed using a translational lead-screw (lead-screw 2 in Fig. \ref{fig: RDO}B). However, to avoid the complexity of also displacing the $y-$stage motor (motor 2 in Fig. \ref{fig: RDO}B) with the change in displacement of the $y-$stage, we implement a a bevel gear mechanism with a square shaft and spur gear transmission. In this system, motor 2 rotates a square shaft via a spur gear system. The square shaft permits passive translational displacements with the $y-$stage, but links the rotational displacements of motor 2 with the rotary displacements of the lead-screw 2. As the spur gear rotates, the square shaft rotates, which in turn rotates lead screw 2, moving the carrier in the $y-$axis direction. 

\subsubsection{Robot Fabrication}\label{subsubsec2}

To ensure MR-conditionality, all associated components are made of plastic or glass. The supporting plates, are printed using poly-lactic acid (PLA) with a fused deposition modeling (FDM) printer (F170, Stratasys, US). The smaller custom designed parts, such as the bevel gears, timing belt pulleys, pneumatic motors, etc., are printed using photosensitive resins with a stereolithography printer (Form 3B+, FormLabs, USA). The commercially available parts, such as carbon fiber rods (which serve as the linear rails) and plastic bearings, are selected to ensure they are all MR-conditional.
Using the compact robot design described in the previous section, along with non-metallic materials, the overall dimensions of the robot is $210\times176\times99$ mm$^3$ with a total weight of $860$ g, lighter than the similar 4-DoF MR-conditional body-mounted robot presented in \cite{li2020body} (weighing $1.5$ kg ). Note that the light weight feature promotes potential clinical adoption for general percutaneous abdomen interventions in pediatric patients, such as  liver biopsy \cite{dezsofi2015liver} or kidney biopsy \cite{gjerstad2023kidney}. 

\subsubsection{Mechatronic Hardware}\label{subsubsec3}

The mechatronic hardware is divided into three domains: 1) electronic, 2) pneumatic, and 3) fiber optic. In the electronic domain, a control interface developed in MATLAB converts high level, surgeon-in-loop commands (entry position and target position) into low level commands (target axes positions) based on inverse kinematics and coordinate frame registration. The high level commands ensure the motor position commands and sequences avoid binding of the spherical joints, which will be further detailed in robot modeling section. The low level commands are communicated to a motion controller (DMC-4163, Galil, USA) through a Local Area Network (LAN) connection. The motion controller is used to (i) send voltage signals to a custom designed valve drive circuit that activates the 3-way 3-position solenoid valves (6425K18, McMaster, USA), (ii) receive feedback signals from the axes' MR-conditional quadrature encoders, and (iii) monitor the four limit switch optic fiber receivers. The motion controller, valve drive circuit, and the solenoid valves are all housed within a custom designed control box. 

In the pneumatic domain, the solenoid valves direct flow from a pneumatic pump to the pneumatic motor. Transmission of the pneumatic fluid is facilitated by 1/4" pneumatic transmission lines (PTL) (5648K74, McMaster, USA). The PTL is connected to the control box and the robot by custom designed connectors that use o-ring style static seals, discussed in \cite{gunderman2023stereotactic}. The PTL and its connectors are fed through the wave-guide between the MR-control room and the MR-imaging suite to decouple the electronic domain from the MRI. The pneumatic motors transfer the angular momentum of the pneumatic flow into rotary actuation applied to the translational lead-screws. In a similar manner, the fiber optic domain transfers light to the fiber optic limit switches to monitor the limits of the range of motion of the translational lead-screws. 

\begin{figure}[t]
    \begin{center}
    \includegraphics[width=0.68\textwidth]{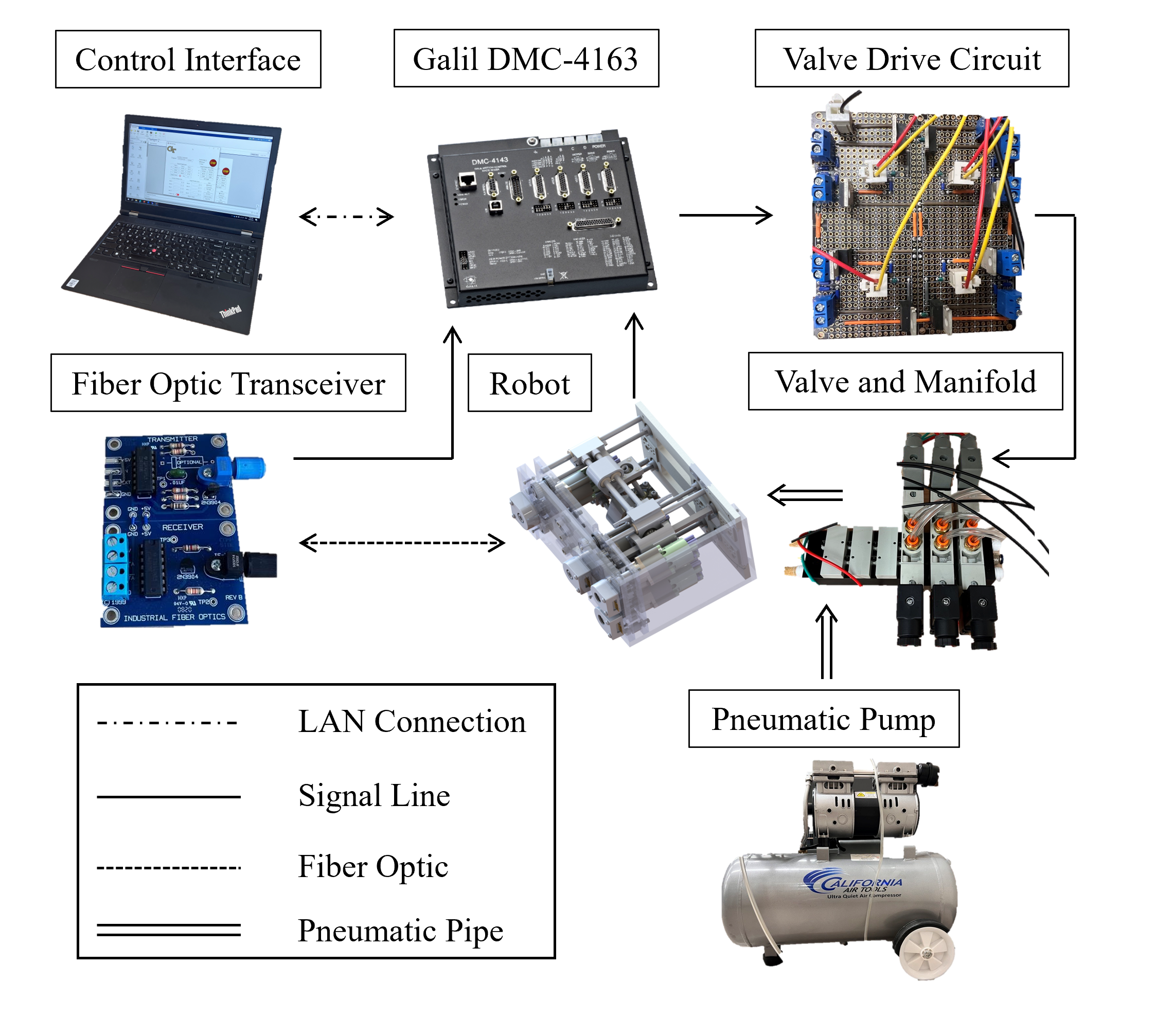}
    \end{center}
    \vspace{-2mm}
    \caption{ 
    The mechatronic hardware system used for controlling the robot. All components. other than the robot. are placed outside of the MR room for safety.}
    \label{fig: MS}
\end{figure}

\subsubsection{Control Strategy}\label{subsubsec4}

In this work, a bang-bang control strategy is implemented on the Galil motion controller based on low level commands (target axes positions) and the axes states. This strategy is used in lieu of a standard PID control algorithm due to the simple, but robust mode of operation of the 3-way 3-position solenoid valves. Although more complicated control algorithms can be used to control the motor for dynamic conditions \cite{gunderman2023Model}, this control strategy is sufficient for set-point tracking in this clinical application, providing a typical carriage accuracy of $<0.5$mm. This level of accuracy is obtainable through the high reduction ratio in our driving system, resulting in a highly damped system and a low carriage translational speed ($<5$ mm/s). This low speed helps minimize carriage overshoot caused by its inertia and any remaining compressed air in the PTL \cite{gunderman2023Model} after the carriage position error is within the threshold ($0.3$mm for $x-$ axis and $0.6mm$ for $y-$ axis), and the pneumatic valves are turned off. 

\subsection{Robot Modeling}\label{subsec2}

\subsubsection{Robot Kinematics}\label{subsubsec5}

The robot inverse kinematics is developed to obtain the desired upper carriage,  $P_u = (x_u, y_u, z_u)$, and lower carriage, $P_l = (x_l, y_l, z_l)$, positions in the robot frame $O_{xyz}$  based on  the desired entry point, $P_e = (x_e, y_e, z_e)$, and target point, $P_t = (x_t, y_t, z_t)$, inputs, as shown in Fig. \ref{fig: WSA}.  Note that the entry point and target point coordinates are obtained from the MR coordinate system and are transformed to the robot coordinate frame using rigid-point  registration between the MR coordinate frame and robot frame. Based on the geometric relationship, the desired position of the carriages can be calculated using the following equations: 
\begin{equation}
    \label{eq: IK}
    \left\{
    \begin{aligned}
        x_u = (z_u - z_t) / (z_e - z_t) \cdot (x_e - x_t) + x_t\\
        y_u = (z_u - z_t) / (z_e - z_t) \cdot (y_e - y_t) + y_t\\
        x_l = (z_l - z_t) / (z_e - z_t) \cdot (x_e - x_t) + x_t\\
        y_l = (z_l - z_t) / (z_e - z_t) \cdot (y_e - y_t) + y_t
    \end{aligned}
    \right.
\end{equation}
where $z_u = -36.5$mm and $z_l = -82.2$mm are the $z-$axis coordinates for two spherical bearings in the robot frame, which are defined during the robot hardware design process.

\subsubsection{Workspace Analysis and Sequential Moving Strategy}\label{subsubsec6}

\begin{figure}[t]
    \begin{center}
    \includegraphics[width=0.68\textwidth]{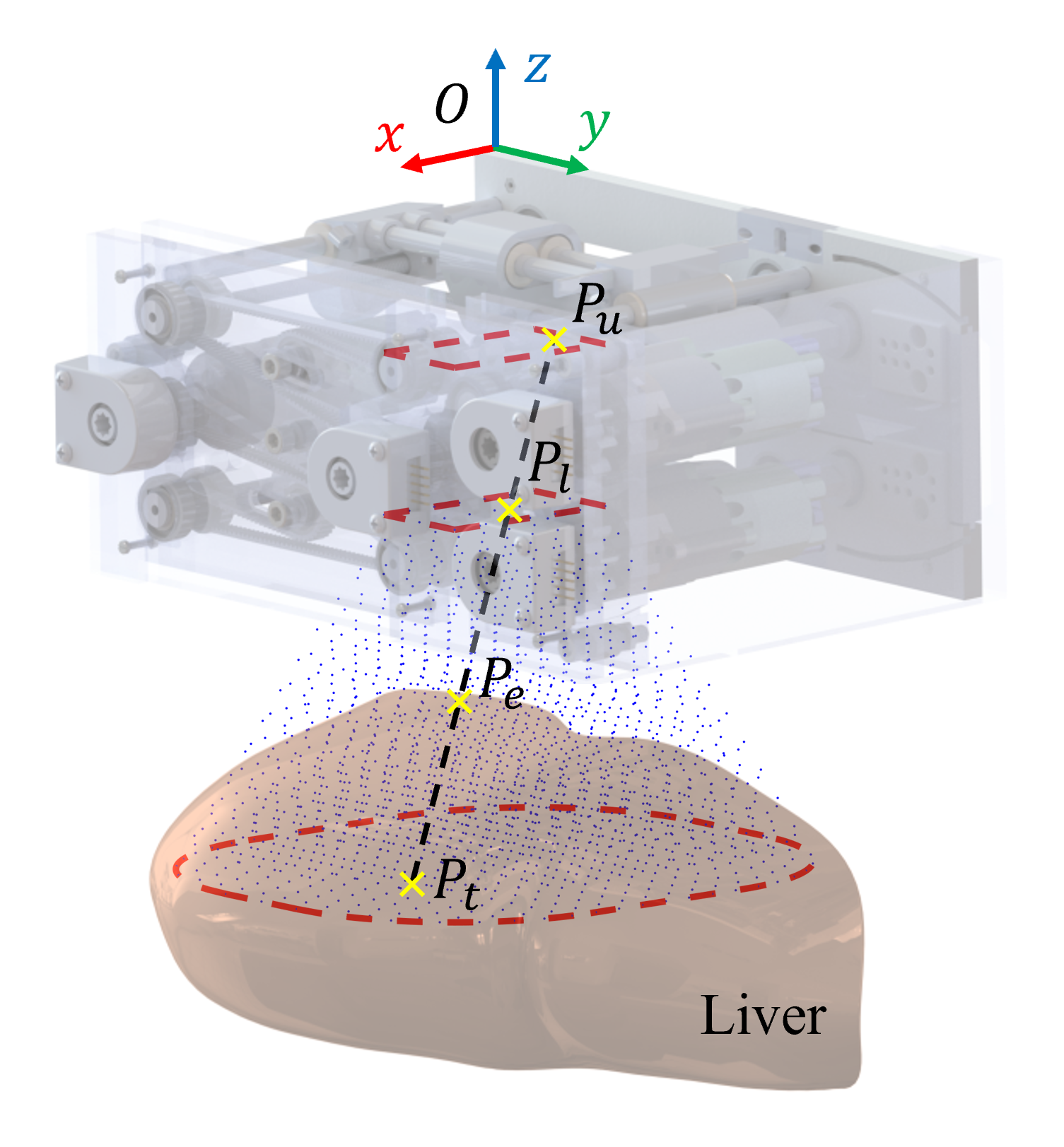}
    \end{center}
    \vspace{-2mm}
    \caption{
    The desired carriage positions $P_u$ and $P_l$ are obtained with the entry point $P_e$ and target point $P_t$ via inverse kinematics. The forward kinematics determines the reachable workspace of the robot, which is a frustum denoted by the blue point cloud superimposed on the liver. Note that the height of the frustum is only limited by the needle length.}
    \label{fig: WSA}
\end{figure}

The robot workspace is dominated by two factors: the feasible travel distance  of the carriages, defined by a $55$ mm by $30$ mm rectangle, and the joint limit of the spherical joints on the carriages, which have a maximum angle of inclination, $\theta$, of $30^\circ$. These constraints result in reachable workspace in the shape of a frustum, as indicated by the point cloud shown in Fig. \ref{fig: WSA}. To assess the liver volume encompassed by the robot's workspace, the workspace point cloud is superimposed onto a liver model. The liver model is appropriately offset from the workspace by a distance of $25$ mm, corresponding to the thickness of the abdominal wall \cite{su2019skin}. Subsequently, the intersection of the workspace with the liver body size is quantified, revealing that the reachable region within the liver occupies 70\% of the liver volume ($1147$ml \cite{urata2000standard}). 

In addition to analyzing the workspace for the robot, a sequential moving strategy is developed that prevents violation of the physical constraints mentioned above during robot motion. This strategy is implemented to ensure that the angle of inclination of the insertion vector that connects both upper and lower carriages is within the $30^\circ$ operational range. Additionally, during the practical implementation of the robot, we noticed that only one pneumatic motor can be effectively actuated at any given time due to the limited volumetric capacity and recharge rate of the compressor. To overcome these challenges, a sequential moving strategy is proposed in Algorithm \ref{alg:cap}. This algorithm coordinates the independent movement of each of the four axes in sequence to ensure the incline angle of the guide stays within its limit. The carriages are sequentially moved using the low-level motion controller with the bang-bang control strategy along the $x$- and $y$-directions in every iteration of the while loop. During each iteration, the carriage with a larger absolute position error is chosen to be moved by a maximum amount of $5$ mm towards the target position. Thus, Algorithm \ref{alg:cap} ensures that the carriages are safely guided to the desired position without violating hardware constraints. It should be noted that a larger compressor will be used in future work to enable simultaneous multi-axis control. 

\begin{algorithm}
\centering
\caption{Sequential Moving Strategy}
\label{alg:cap}
\begin{algorithmic}
\State $i \gets 0$
\While{robot is not in target position}
    \State $e_{a1} \gets \text{position\ error\ of\ axis\ 1}$
    \State $e_{a2} \gets \text{position\ error\ of\ axis\ 2}$
    \State $e_{a3} \gets \text{position\ error\ of\ axis\ 3}$
    \State $e_{a4} \gets \text{position\ error\ of\ axis\ 4}$
    \If{$i = 0$}
        \State $d \gets min(max(|e_{a1}|, |e_{a3}|), 5)$
        \If{$|e_{a1}| > |e_{a3}|$}
            \State move axis 1 by $d\times\sign(e_{a1})$ mm
        \Else
            \State move axis 3 by $d\times\sign(e_{a3})$ mm
        \EndIf
    \Else
        \State $d \gets min(max(|e_{a2}|, |e_{a4}|), 5)$
        \If{$|e_{a2}| > |e_{a4}|$}
            \State move axis 2 by $d\times\sign(e_{a2})$ mm
        \Else
            \State move axis 4 by $d\times\sign(e_{a4})$ mm
        \EndIf
    \EndIf
    \State $i = 1 - i$
\EndWhile
\end{algorithmic}
\end{algorithm}

\subsection{Clinical Workflow }\label{subsec0}
In this work, our long term goal is to develop a robot capable of providing position and orientation control of our step insertion unit \cite{musa2021respiratory} that enables active needle insertion within a closed-bore MRI under real-time MRI guidance. Based on the consultation with the clinical support, here we provide the surgical workflow: 

\begin{enumerate}[label={\arabic*)}]
    \item The patient is brought into the MRI suite, positioned on the table, and anesthesia is started.
    \item The robotic platform is placed on the region of interest and secured to the patient using straps. MWA needle will be placed and secured in the automatic insertion module.
    \item MRI will be performed to localize the targets and prepare path planning in the navigation system. The interventional radiologist selects the target points and skin entry points. The robot coordinates are registered to the MR coordinates based on the MRI tracking coils embedded in the robot.
    \item The robot is then commanded to move and orient the needle along the prescribed path to treat the selected target via real-time MRI tracking coil feedback. The navigation software will display the projected needle insertion path into the tissue across the field of view. The physician verifies the planned path and adjusts as needed. 
    \item The robotic needle insertion module will stepwise deploy the needle based on the respiration cycle. During the insertion process, real-time MRI will be performed to track the needle position and target location to ensure safety, and allow the clinician to monitor/correct the treatment as needed.
    \item Once the needle is placed, a high-resolution MRI scan is performed to confirm the ablation needle location.
    \item MWA procedure is then performed with intraoperative thermometry monitoring.
    \item Steps 4-6 are repeated to treat other targets if needed.
    \item Post-procedure MRI will be performed to assess the effectiveness of the treatment.
\end{enumerate}

Prior to the needle insertion process, the clinician will confirm the robot location and the treatment trajectory. This human-in-loop control will allow us to safely deploy the needle without damaging the critical region. Also note in this early stage, we only present a prototype to facilitate needle-guide in a closed-bore MRI scanner without automatic needle insertion module. 

\section{Results}\label{sec3}
\subsection{Free-Space Axis Position Accuracy Validation}\label{subsec3}

\begin{figure}[t]
    \begin{center}
    \includegraphics[width=0.68\textwidth]{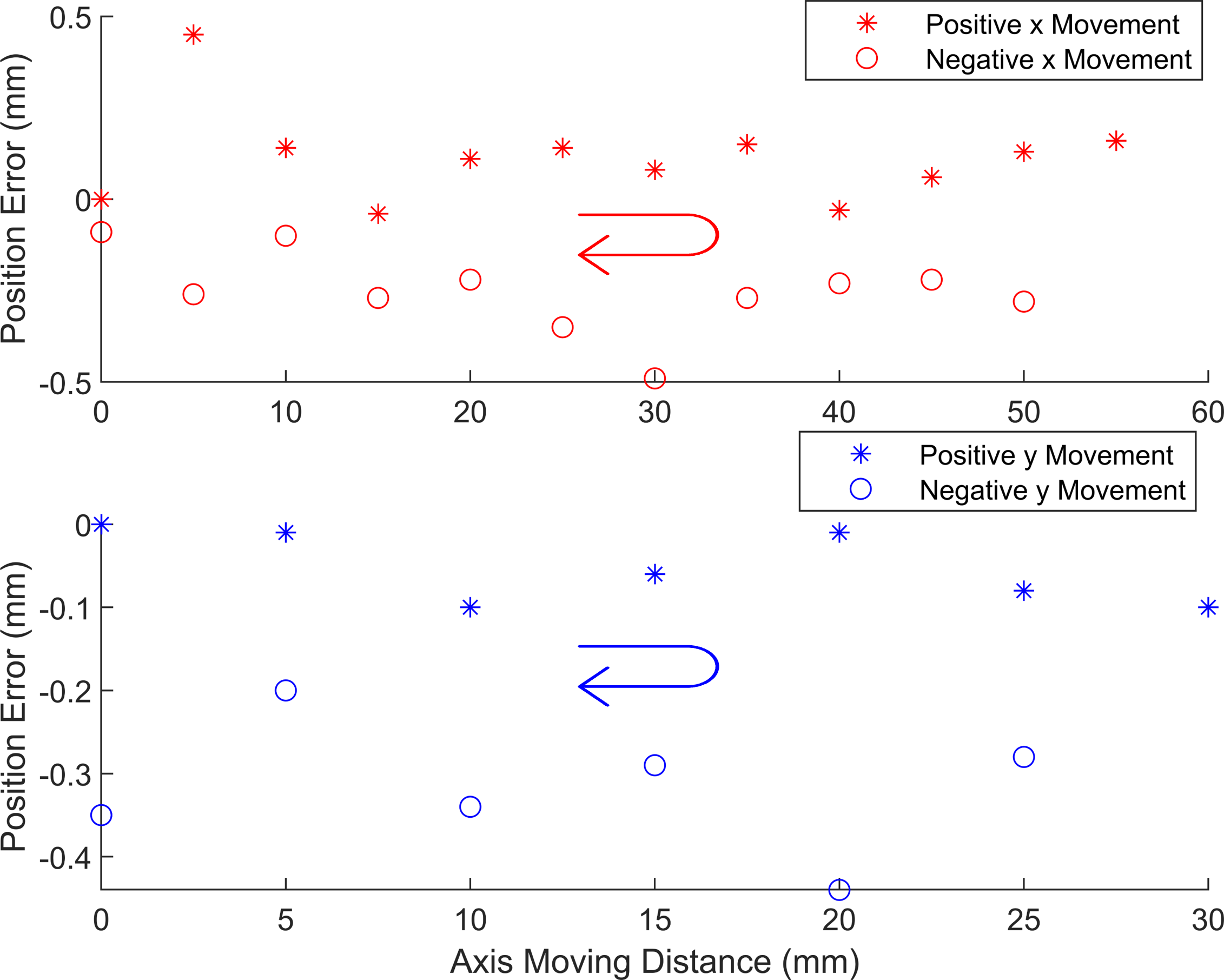}
    \end{center}
    \vspace{-2mm}
    \caption{The carriage was translated along the $x$- (red) and $y$- (blue) axes, respectively. The position error exhibit variations across different axes and directions, but both are within sub-mm error. }
    \label{fig: SAA}
\end{figure}

Prior to the system level evaluations, positioning accuracy of the independent axes of the Cartesian stages were evaluated to quantify the error of the transmission system assembly and motor control performance. To fully validate its position accuracy, the carriage was first driven forward from its home limit position along the positive axis direction to the maximum limit position in 5 mm increments. Subsequently, it was moved along the negative axis direction back to the home limit position in -5 mm increments. At each increment, the actual carriage position relative to the robot frame was recorded using a standard vernier caliper (500-196-30, Mitutoyo, Japan) with a resolution of $0.025$ mm. Both the $x-$ and $y-$ stages were actuated to their corresponding limits, providing a total of 36 data points as shown in Fig. \ref{fig: SAA}. The mean absolute position error was $0.19\pm0.13$ mm for the $x-$stage and $0.17\pm0.15$ mm for the $y-$stage.

Notice that the position error behaves differently between different axes and directions, as indicated in Fig. \ref{fig: SAA}. Note that while the mean error is calculated as the absolute position error, Fig. \ref{fig: SAA} depicts the signed error to highlight the differing axis behavior. This difference is primarily caused by differing carriage translational velocities, likely attributed to the manufacturing tolerances in the additive manufactured system, which impacts the bang-bang control strategy's performance. For example, during the experiment, it was observed that the $x-$stage of the upper Cartesian platform has a relatively uniform velocity in opposing directions. Conversely, the $y-$stage of the upper Cartesian stage requires $60$s to transition from $0$ mm to $30$ mm, and $35$ seconds to return back from $30$ mm back to $0$ mm. Clearly, displacements in the negative $y-$direction for the corresponding stage have higher translational speeds, causing the system to be prone to overshoot, resulting in larger position errors. This is supported by the data in Fig. \ref{fig: SAA}, where negative displacements in the $y-$axis direction indicate a larger mean absolute error value ($0.32\pm0.08$ mm) compared to negative displacements ($0.05\pm0.04$ mm). However, it should be noted that even when considering the worst case scenario (two carriages with a $0.5$ mm position deviation towards opposite directions along the diagonal), the needle tip position error on a targeting plane with a depth of $100$ mm would only be $3.8$ mm, which is relatively small compared to the clinically-approved RFA HCC tumor size of $2$-$3$ cm, and the typical ablation volume of 4 cm \cite{rhim2010radiofrequency}. 

\subsection{Free-Space Robot Targeting Accuracy Validation}\label{subsec4}

\begin{figure}[t]
    \begin{center}
    \includegraphics[width=0.68\textwidth]{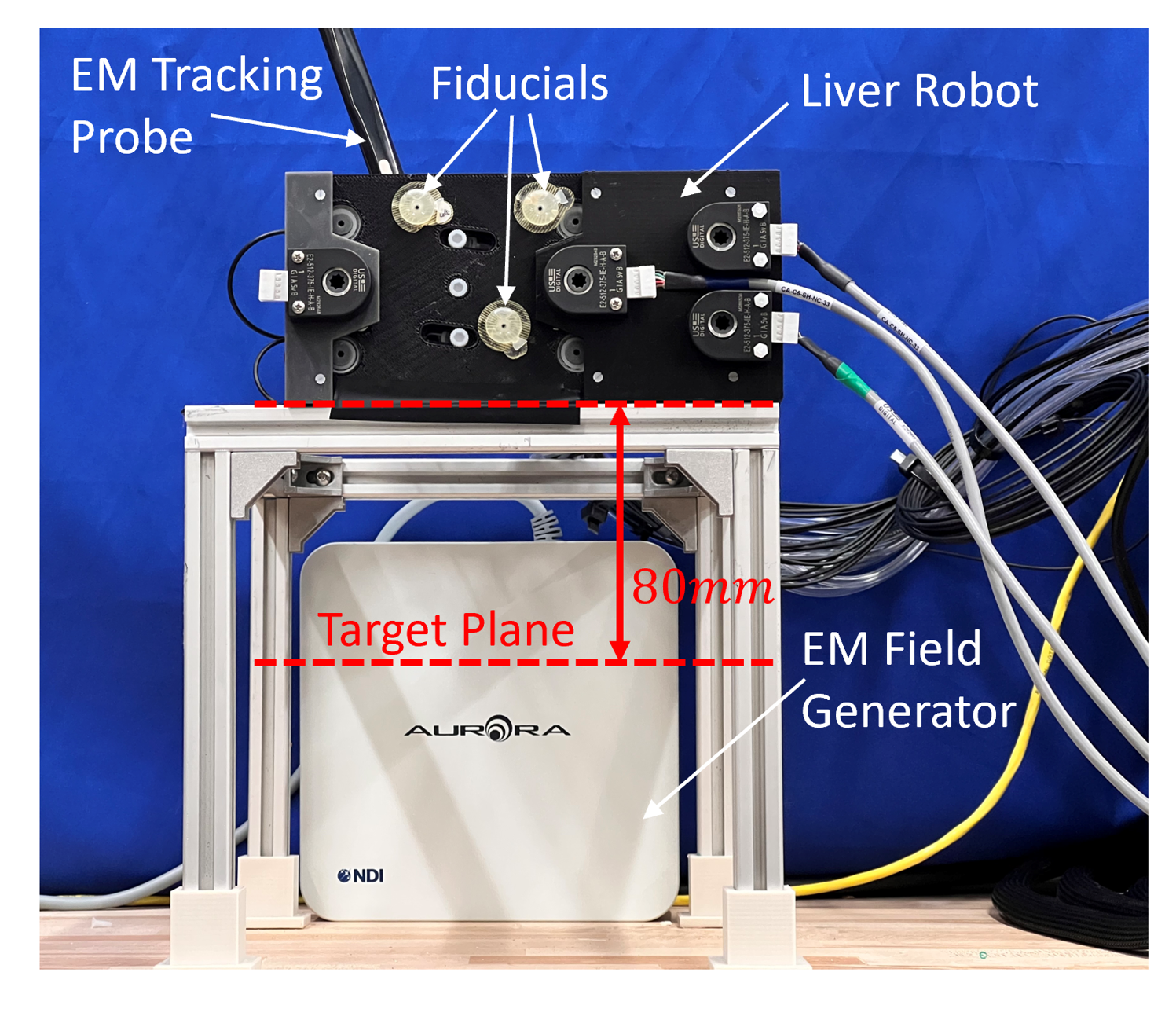}
    \end{center}
    \vspace{-2mm}
    \caption{The robot is firmly secured on an aluminum extrusion frame for targeting characterization. The EM field generator is mounted under the robot to accurately measure the needle tip position with an EM tracking probe. Three fiducials are attached to the front supporting plate of the robot, which will be utilized for registration during MRI-guided validations. }
    \label{fig: FS}
\end{figure}

\begin{figure}[t]
    \begin{center}
    \includegraphics[width=0.68\textwidth]{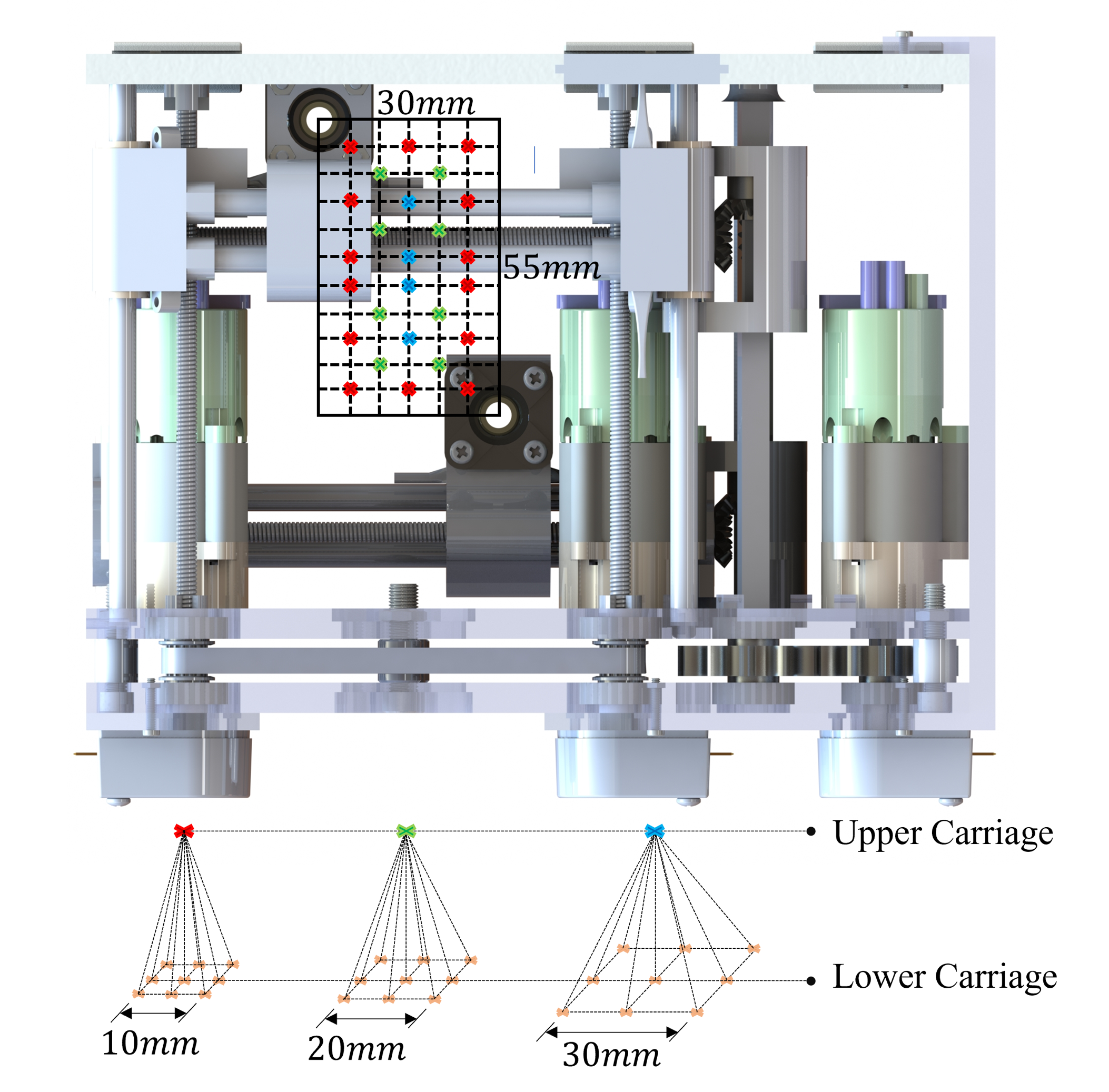}
    \end{center}
    \vspace{-2mm}
    \caption{Framework for sampling points within the robot workspace for the free-space accuracy test. The relation between lower carriage grid layout dimension and upper carriage position is indicated by different colors. This resulted in a total of 234 data points. }
    \label{fig: DP}
\end{figure}
A free-space bench-top experimental validation was performed to test the robot's guiding accuracy for needle placement,  as seen in Fig. \ref{fig: FS}. For this experiment, 234 targets were selected at an insertion depth of $80$mm to ensure a well-distributed coverage of carriage positions. The target poses consist of 26 points that were chosen in the upper carriage and each was grouped with 9 points in the lower carriage in a 3-by-3 grid layout, as depicted in Fig. \ref{fig: DP}. The grid layouts were designed to examine the robot accuracy performance at different angles of inclination and positions. 

During the experiment, the measurement was performed using an EM tracking system (Aurora,  NDI Medical Inc.), which has a measurement error of $0.5$ mm. To obtain the needle insertion vector, the EM tracking probe was inserted into the needle guide twice, once from the spherical bearing on the upper carriage towards the lower carriage, and a second time from the lower carriage to the top one. This process yielded two measured probe tip positions, which were subsequently transformed into the robot frame based on the rigid-point registration. These points were used to produce the position and orientation of the needle insertion vector. This vector was then extended to project a point onto the target plane that was at a depth of $80$ mm, indicating the needle tip position if inserted, as shown in Fig. \ref{fig: FS}. 

\begin{figure}[t]
    \begin{center}
    \includegraphics[width=0.68\textwidth]{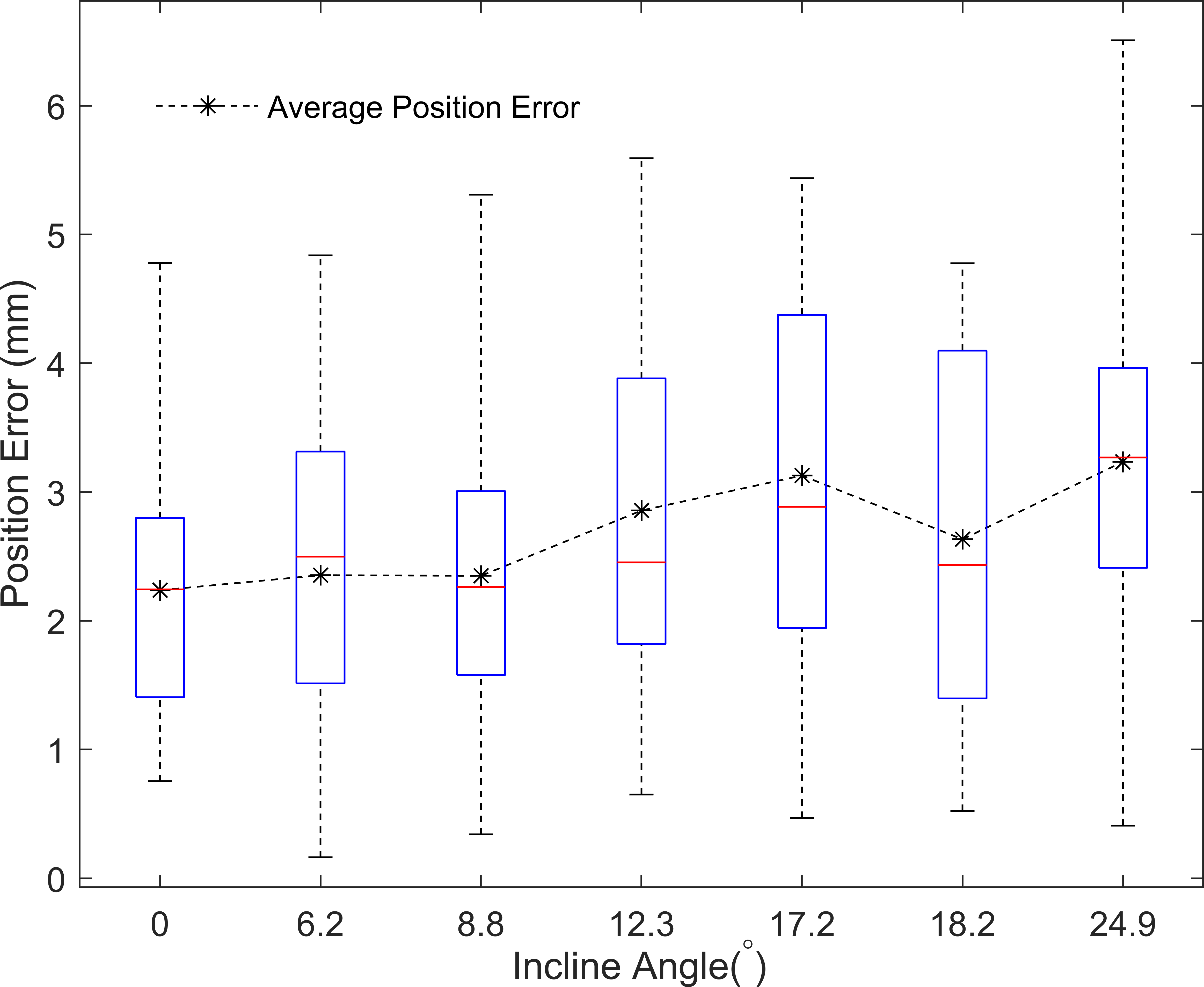}
    \end{center}
    \vspace{-2mm}
    \caption{Position error vs. angle of incline. The position error is increases with increasing incline angle due to the increasing needle travel distance. }
    \label{fig: FSR}
\end{figure}

The position error of the needle tip was defined as the Euclidean distance between the target location and the intersection point of the needle guide pointing direction and the target plane. The average position error across all targets was found to be $2.6\pm1.3$ mm. The orientation error was defined as the angle difference between desired and measured needle insertion vector, calculated using their dot product. The average measured value of $3.9\pm1.2^\circ$. The 234 data points were divided into 7 groups by their incline angle in order to provide a more detailed quantitative assessment of the robot's targeting accuracy.  As shown in the Fig. \ref{fig: FSR}, the average position error increased with the incline angle due to the longer travel distance between the spherical bearings and the target plane, indicating the inherent variance in the accuracy performance of the robot. 

\subsection{MRI Phantom Validation}\label{subsec5}

\begin{figure}[t]
    \begin{center}
    \includegraphics[width=0.68\textwidth]{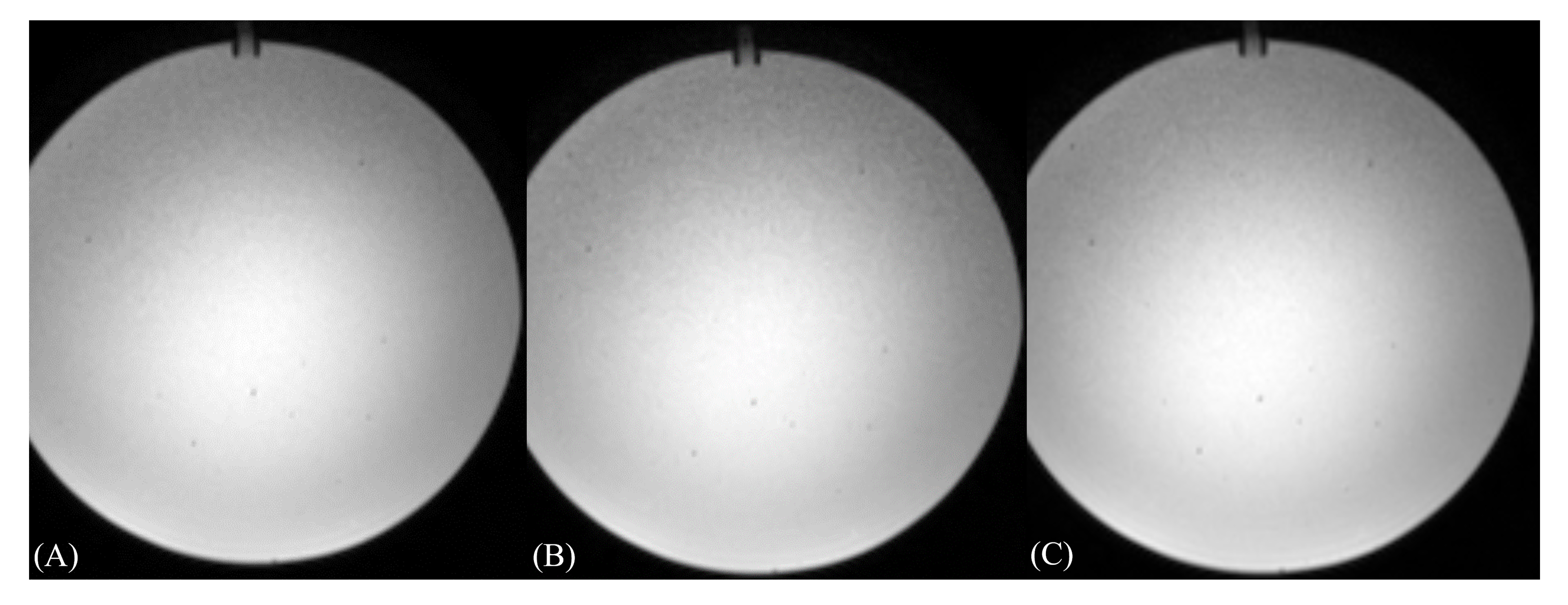}
    \end{center}
    \vspace{-2mm}
    \caption{The MR-conditionality validation result indicates that the MRI image of a water phantom varies insignificantly under different robot statuses: (A) image without robot; (B) image with robot on; (C) image with robot off.}
    \label{fig: MRICV}
\end{figure}

MR-conditionality evaluations were performed in a 3T Philips Ingenia Elition MRI scanner using 3D spoiled gradient-echo imaging. The robot was placed $20$ mm away from a water-phantom. The MRI image (FOV: $224\times224\times45 $ mm$^3$, $1\times1\times5$ mm$^3$ resolution, TR/TE = $13/2$ ms) of the phantom was acquired under three scenarios: no robot, robot off while in the scanner, and robot on while in the scanner. No obvious differences were observed in the images between the different scenarios, as shown in Fig. \ref{fig: MRICV}, demonstrating the robot's MR-conditionality with MR imaging. Additionally, the Signal-to-Noise Ratio (SNR) \cite{kaufman1989measuring} of three images in Fig. \ref{fig: MRICV} was calculated to be $35.7$ dB (without robot), $35.5$ dB (with robot off) and $34.2$ dB (with robot on), suggesting a negligible impact on image quality due to the presence of the robot. 

\begin{figure}[t]
    \begin{center}
    \includegraphics[width=0.68\textwidth]{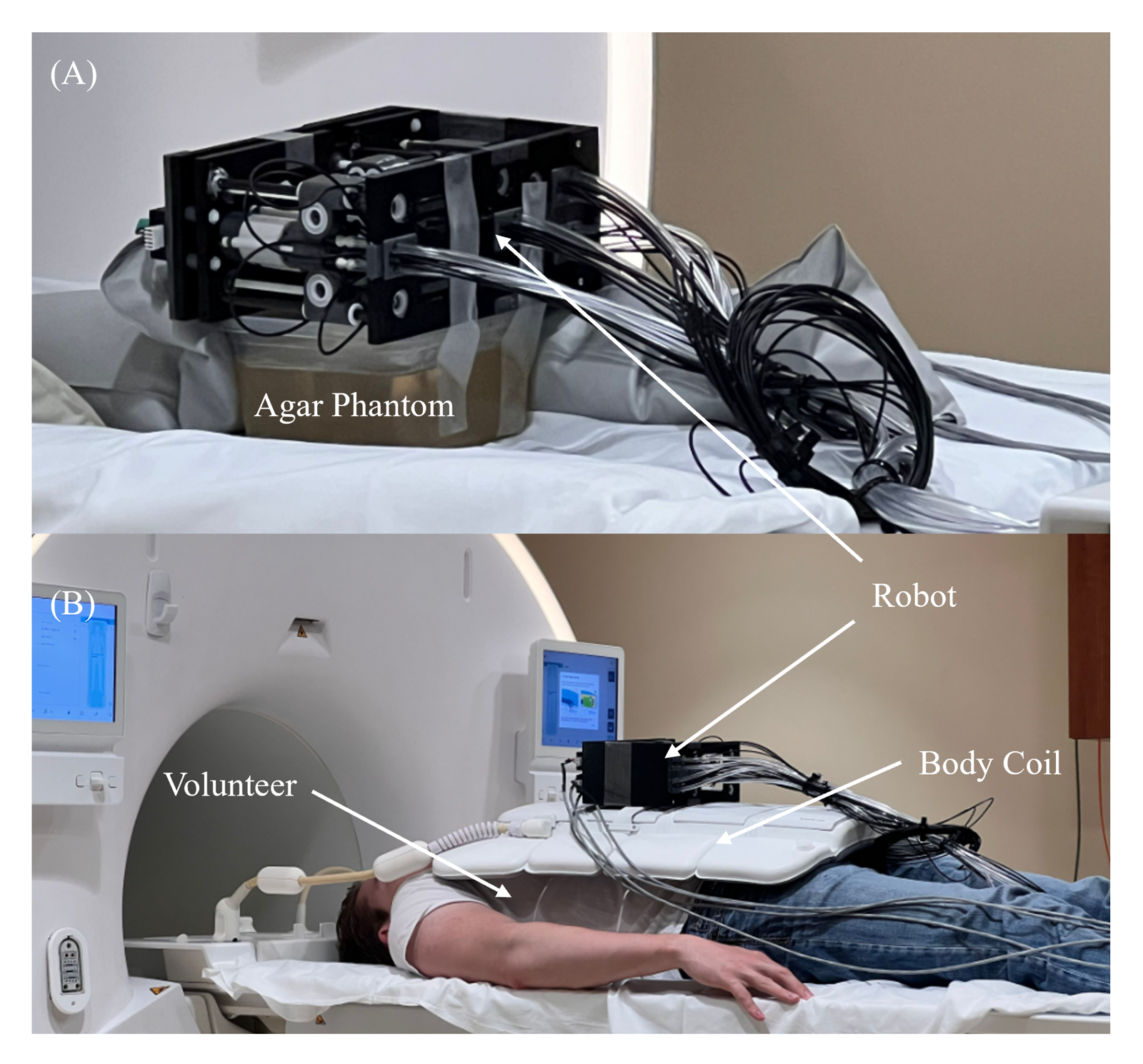}
    \end{center}
    \vspace{-2mm}
    \caption{(A) MRI phantom insertion experiment with the robot mounted above the agar phantom. The robot is restrained to the agar phantom using adhesive medical tape. (B) A depiction of the proposed system resting on a patient with the body coils. Note that the robot can be used within a standard MRI bore. }
    \label{fig: MRIS}
\end{figure}

\begin{figure}[t]
    \begin{center}
    \includegraphics[width=0.68\textwidth]{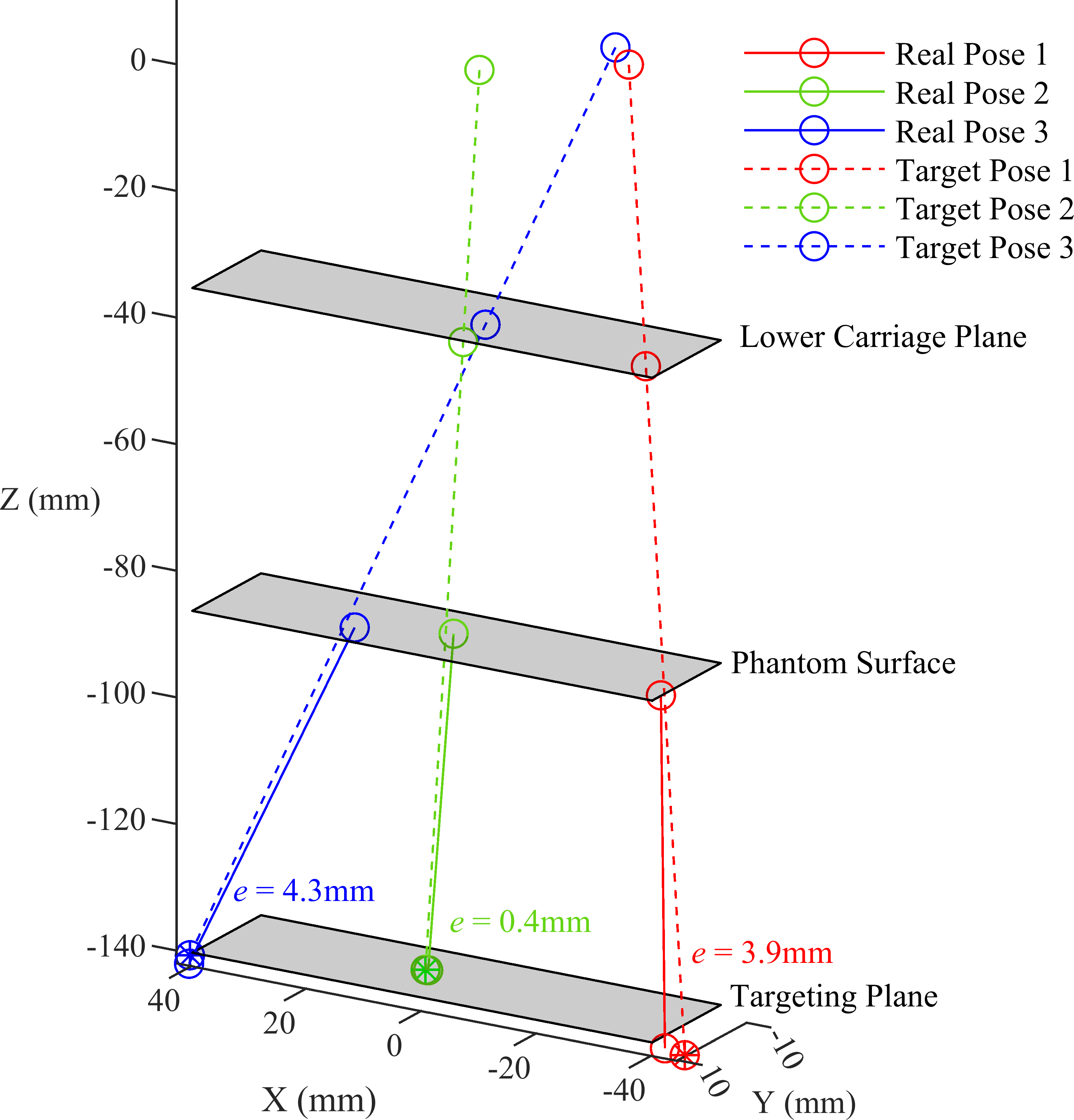}
    \end{center}
    \vspace{-2mm}
    \caption{Three insertions were implemented during the experiment and the corresponding needle tip position error $e$ was measured using 3D MR images. }
    \label{fig: MRIR}
\end{figure}

To simulate targeting in biological tissue inside the MR environment, a 10\% by weight Knox$^{\text{TM}}$ (Kraft Foods Global, Inc., USA) gelatin phantom insertion experiment was conducted in the same MRI scanner, as shown in Fig. \ref{fig: MRIS}A. Note that while the experiment performed in a phantom study, Fig. \ref{fig: MRIS}B indicates that the robot can be used within the MRI bore with a large volunteer (113 kgs). It also indicates that the robot meets the size constraint imposed by the body coil. However, we want to highlight that the body coil depicted in Fig. \ref{fig: MRIS}B is not suitable for practical use as it potentially interfaces with the needle entry path. The development of a custom-designed body-coil, complete with a detailed mounting mechanism between the robot and the coil opening, will be investigated in our future work. This will be critical in facilitating the use of the robot in commercial MRI scanners in the long-term. In the current configuration, straps can be used to mount the robot to the patient. For the phantom experiment, a total of three insertions were performed, as seen in Fig. \ref{fig: MRIR}. For each insertion, a random robot kinematic pose was specified and the robot was commanded to move to that pose. The target position was then defined as the needle tip position on the targeting plane as a result of the needle vector defined by the desired kinematic pose of the needle guide. A needle was then manually inserted to a targeting plane located at a depth of $105$ mm from the robot base within the phantom. The real needle insertion vector and tip position were then measured using 3D spoiled gradient-echo imaging (FOV: $192\times192\times192 $ mm$^3$, $1\times1\times1$ mm$^3$ resolution, TR/TE = $13/4$ ms) and converted to the robot frame using coordinate registration performed with three MR-visible fiducials mounted on robot supporting plate as shown in Fig. \ref{fig: FS}. The position error of the needle tip was defined as the Euclidean distance between the target location identified by the kinematic model and the measured location of the needle tip in the robot frame. The orientation error was defined as the angle difference between desired and measured needle insertion vectors in the robot frame, calculated using their dot product. The results indicated a tip position error of $2.9\pm2.1$ mm and an orientation angle error of $2.1\pm1.4^\circ$ between them, where both errors are defined in the same manner as in the free-space test. 

\section{Discussion}

In this paper, a novel design of a 4-DoF MR-conditional robot for assisted MR-guided needle insertion was presented. We demonstrated its capability for needle guidance in free-space and within the MRI bore. The robot consists of two stacked Cartesian stages that work in tandem to manipulate a needle guide. Each stage provides two degrees of freedom, equipping the needle guide with a total of 4 controllable degrees of freedom. The insertion is controlled manually using a surgeon-in-loop approach, while each controllable axis of the Cartesian stage is actuated using an MR-safe pneumatic turbine with a large gear reduction. All electronics used for control are decoupled from the MRI via pneumatic and fiber optic transmission modalities. This design is compact and capable of fitting within a standard MRI-bore even with large (113 kg) patients.

The system was first evaluated at the axis level. Despite the high latency of the pneumatic actuating system due to PTL dynamics, and the simplicity of the bang-bang control algorithm, the system yielded an average axis movement accuracy of $0.18\pm0.13$ mm, primarily due to the large reduction and damping associated with the plastic gearbox and translational lead-screws. Even with this associated error, the intrinsic design of the parallel manipulator promotes error mitigation when compared to manipulators due to the non-accumulating joint error, which is a beneficial feature of this robotic design \cite{simaan1999analysis}. After evaluation on the axis level, a system level evaluations were performed. Despite the needle length, which amplifies the axis-level error, the needle tip error remained below 5 mm ($2.6\pm1.3$ mm at an insertion depth of 80 mm) in both the free-space and MRI phantom experiments. Additionally, the angular error remained below 5°. It should be noted that the maximum observed axis error was 0.5 mm. If the axes exhibited an error with the worst case scenario (two carriages with a 0.5 mm position deviation towards opposite directions along the diagonal), this would suggest an angular error of 1.77°, which is only a fraction of the free-space (3.9$\pm{}$1.2°) and MRI (2.1$\pm{}$1.4°) experimental angular error. This suggests there are other errors beyond those defined by the axis error. Likely causes include the registration error of the system, fabrication errors associated with angular play in the spherical joint and the linear rail, and free-space between the needle guide and the needle. 

The robot in this study serves as a proof-of-concept for the MR-guided treatment of HCC and other abdominal interventions. Its light-weight design and MR-guided implementation also promotes its use in potential pediatric applications. While the needle tip error below 4mm meets the requirements for most clinical scenarios, it is noteworthy that the robot workspace is constrained by a maximum insertion angle of $30^\circ$ and hence leads to a limited selection of possible needle paths towards target. This limitation is primarily imposed by a relatively large spacing between the two Cartesian stages of the robot and the limited incline angle provided by the spherical bearings. Another limitation of our study is the cost associated with MRI guided experiments, which precludes the collection of statistically significant results in this early stage. While our goal is to validate the proof-of-concept here, our future work will focus on improving the robot targeting accuracy, enhancing the workspace, optimizing the robot dimension, and extensive validations inside MRI environments. To achieve these goals, we plan to implement dynamic MRI image feedback \cite{gunderman2023non}, replace the pneumatic valves with directional-proportional control valves \cite{gunderman2023Model}, and utilize fabrication methods with higher accuracy. These additions will enable us to further improve the accuracy of needle insertions and increase the robot's operational capabilities within the MRI environment. Finally, we will investigate the applicability in path-planning for the needle while avoiding sensitive abdominal structures, similar to our prior work in neurosurgical interventions \cite{Huang2023CTRDesign}.

\section*{Declarations}
\subsection*{Conflicts of interests}
The authors declare that they have no known competing financial interests or personal relationships that could have appeared to influence the work reported in this paper.

\bibliography{reference} 

\end{document}